\documentclass{article}

\usepackage[preprint]{neurips_2026}

\usepackage[T1]{fontenc}
\usepackage{url}
\usepackage{booktabs}
\usepackage{amsfonts}
\usepackage{nicefrac}
\usepackage{microtype}
\usepackage{xcolor}
\usepackage{graphicx}
\usepackage{amsmath}
\usepackage{amssymb}
\usepackage{array}
\usepackage{float}
\usepackage{tikz}
\usetikzlibrary{positioning,arrows.meta,fit}
\usepackage{hyperref}
\hypersetup{colorlinks=true, allcolors=[rgb]{0.10,0.24,0.52}}
\title{VAmoS Part Deux: Harder, More Realistic Voice-Agent Simulation}

\author{%
  \textbf{Joshua Meyer} \quad
  \textbf{Sahar Shayegan} \quad
  \textbf{Ritiz Tambi} \quad
  \textbf{Ali Khan} \\
  \textbf{Sun Kim} \quad
  \textbf{Victor Shih} \quad
  \textbf{Mehdi Jamei} \quad
  \textbf{Andi Partovi} \\
  \normalfont Veris AI \\
  \normalfont\texttt{joshua@veris.ai}
}

\begin{document}
\maketitle

\begin{abstract}
Voice agents in production must handle several requests, background speech,
and customers who lose patience. We introduce VAmoS Energy, a benchmark that
combines these challenges in 100 calls about utility billing and payment
assistance. Each caller makes two to four requests. The agent has sixteen
tools backed by a stateful Stripe billing twin and the Apache Fineract loan
engine, with account access blocked until caller verification succeeds.
The tasks use public household electricity data and a policy based on
Pennsylvania's residential billing rules. An LLM-as-a-verifier checks the
agent's actions and spoken figures against explicit requirements. On a
calibration run, it agrees with a code verifier on
99.1\% of checks. Across
fourteen voice stacks and three repeats per task,
completion ranges from 17.3\% to
44.7\%. Grok Voice leads, and Gemini 3.8 Live and GPT-Live
follow at about the same cost per call. Background television reduces pooled completion from
38.7\% to 8.6\%.
The simulated caller often accepts an incorrect result because it hears the
agent's words but cannot inspect its actions. These findings show why voice
agents need evaluation across the whole call, including what they say, what
they change, and how they handle competing speech.
\end{abstract}

\section{Introduction}
\label{sec:intro}

Calls in production can demand more than a benchmark's single request in a
quiet room. A customer may question a bill, make a payment, and change a
billing preference while a television plays nearby. VAmoS Bench
\citep{meyer2026vamosbench} evaluated voice agents on credit-card support
calls by grading their conversations together with their tool calls and
results. The best voice agent completed 71\% of calls. Its most difficult
scenarios required several steps.

We introduce VAmoS Energy, a benchmark for utility billing and payment
assistance. Each task gives the caller two to four requests. The agent must
verify the caller, consult account records in two stateful services, and
follow a policy based on Pennsylvania's residential utility regulation.
Public household electricity data supplies the usage profiles. We evaluate
fourteen voice stacks with calm and angry callers under
four background conditions. Table~\ref{tab:compare} summarizes the changes
from VAmoS Bench. The completion rates are lower, but the domains and agent
configurations also differ, so this is not a controlled comparison of task
difficulty alone.

\begin{table}[htbp]
\centering\small
\caption{The two VAmoS benchmarks. Observed completion depends on both the
benchmark and the evaluated configurations.}
\label{tab:compare}
\begin{tabular}{@{}p{2.7cm}p{4.3cm}p{5.7cm}@{}}
\toprule
 & VAmoS Bench & VAmoS Energy \\
\midrule
Backend & Seeded PostgreSQL & Stripe twin + Apache Fineract \\
Agent tools & 5 & 16 \\
Verification & Card digits, name, address/phone & Account number, name, second factor \\
Requests per call & One, sometimes multi-step & 2--4 \\
Data & Generated & ResStock usage; Pennsylvania policy \\
Caller personas & Per-scenario style & Calm, angry \\
Background & Clean & Clean, street, caf\'e, TV \\
Grading criteria & Task-specific assertions & Checks derived from account data \\
Domain & Credit-card support & Utility billing and payments \\
Completion & 43.0--71.0\% & 17.3--44.7\% \\
\bottomrule
\end{tabular}
\end{table}

\paragraph{Harder calls.} We build tasks by combining customer requests,
checking that each remains valid after the preceding actions, and using
measured success rates to select difficult combinations
(Section~\ref{sec:exam}). Callers may be angry and may speak over street
noise, caf\'e chatter, or television. Every stack receives the same persona
and background assignment for each task and repeat.

\paragraph{Stateful services and restricted access.} Each call gets its own
copy of a consistently seeded Stripe twin and Apache Fineract instance.
Tool code blocks access to customer records until verification succeeds.
The agent must then choose and carry out actions that satisfy the customer's
request and the utility's policy.

\paragraph{Checking words and actions.} We use an LLM-as-a-verifier to check
specified facts, including amounts, tool results, and the order of consent
and action. The simulated caller is itself a voice agent and can depart from
its instructions. We audit these deviations along with grading and platform
errors, and exclude affected calls from the reported scores
(Section~\ref{sec:validity}). The verifier can inspect both words and actions;
the caller can only hear what the agent says.

\paragraph{Findings.} Agents usually respect the verification gate, but often
fail to make the required account changes. Success rates fall as callers add
requests. Television is the hardest background condition. Repeated trials
produce different outcomes, and 54.2\%
of calls the simulated caller considers successful fail verification.

\section{Related work}
\label{sec:related}

EVA-Bench evaluates task completion from final database state alongside
faithfulness and speech fidelity, and includes accent and noise perturbations
\citep{bogavelli2026evabench}. $\tau$-Voice evaluates full-duplex agents on
stateful tasks and systematically varies noise, accents, and turn-taking
\citep{ray2026tauvoice}. VAmoS Bench grades conversation and execution
together across production voice stacks \citep{meyer2026vamosbench}.
VAmoS Energy combines multiple requests per call, caller disposition, and
background conditions in a utility domain with two vendor services.

$\tau$-bench pairs a simulated user with domain policies and tool APIs; its
reward checks final database state and required response information
\citep{yao2024taubench}. SWE-bench evaluates code changes through repository
tests \citep{jimenez2024swebench}. LLM-as-a-judge supports open-ended
assessment but can show position and verbosity biases
\citep{zheng2023judging}. Our verifier uses explicit task checks while relying
on an LLM to interpret conversational evidence.

Recursive Synthetic Terminal Tasks (RST) grows difficult tasks by extending
solutions, updating instructions and verifiers, and validating each task in
a sandbox before reusing it \citep{li2026rst}. We use a related recursive
construction, adding valid requests to existing calls, then selecting tasks
using measured difficulty. Our admission test uses an executable policy
model; it does not establish solvability through a successful reference
agent. The environment follows the simulation approach of RAISE
\citep{shayegan2025raise}, applied here to real-time voice calls.

\section{The benchmark}
\label{sec:bench}

\subsection{Task, verification, and policy}
\label{sec:agent}

The agent, Rory, answers a residential utility's billing line. It can explain
bills, take payments, arrange instalments, report existing arrangements, and
change billing preferences. Every stack receives the same system prompt and
sixteen tool definitions (Appendix~\ref{app:tools}).

Before accessing an account, the agent asks for the account number, full name,
and one second factor: card last four digits, postal code, or last bill amount.
The verification tool checks both services and returns whether the details
match. On failure it reports the remaining attempts, without revealing account
data or which detail was wrong. Three failures lock further verification.
All account tools refuse access until verification succeeds; transfer to a
human remains available. Later tools use the verified account rather than an
account identifier supplied by the model. The prompt also forbids guessing
which factor failed or offering a corrected account number. The tool prevents
retrieval of those facts, but cannot prevent the model from inventing them.

\paragraph{Policy.} Correct assistance requires more than a successful API
call. The prompt tells the agent which actions it may offer and when it needs
consent. For example, it must quote a payment arrangement before creating it,
confirm the destination email before enabling paperless billing, and report
a declined payment accurately. A promise to pay later must not become a
recorded payment. These requirements determine the task's success checks.

\subsection{Stateful services and public data}
\label{sec:world}

Veris runs two stateful services for each simulated call: a Stripe billing
twin and the actual Apache Fineract loan engine. The same account number
links a Stripe customer to a Fineract client. Invoices and card payments live
in Stripe; payment arrangements are zero-interest Fineract loans. The
services retain changes during the call and enforce their own API rules.
Fineract, for example, rejects future-dated repayments.

We seed both services from one account manifest, read the records back, and
check their consistency before taking a snapshot. Each call starts from a
fresh clone of that snapshot, so its writes cannot affect another call.
The snapshot freezes time at 2026-09-10T00:59:09Z, keeping due dates and
arrears consistent across attempts.

The 24 accounts in the exam use four months of modeled
Pennsylvania household electricity consumption from ResStock 2022.1
\citep{nrel2022resstock}. We aggregate 15-minute readings for June through
September using actual 2018 weather, without scaling, and select profiles
with each month between 150 and 1{,}000\,kWh. The data describes simulated
homes, not customer meter readings. Prices, taxes, and fixed charges are
benchmark assumptions.

\paragraph{Policy grounded in regulation.} Utility regulation constrains
payment assistance and service termination. We used Pennsylvania's Chapter~56
\citep{pacode56} to inform the policy supplied in the system prompt, including
income-dependent arrangements, winter protection, and medical certificates.
The benchmark encodes a fixed, simplified policy together with authored
business rules. Appendix~\ref{app:policy} states the encoded terms and their
limits; they should not be read as a complete statement of current law.

\subsection{Building difficult tasks}
\label{sec:exam}

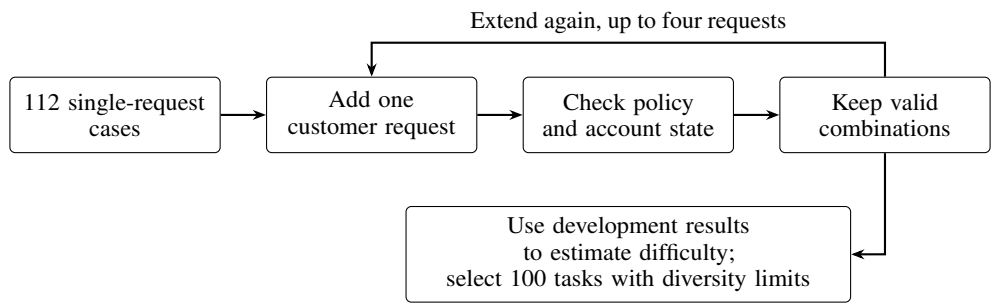
\begin{figure}[t]
\centering
\begin{tikzpicture}[
 box/.style={draw,rounded corners=2pt,align=center,font=\small,text width=2.5cm,minimum height=1cm,inner sep=4pt},
 arr/.style={-{Stealth[length=5pt]},thick}]
\node[box] (seed) {112 single-request\\cases};
\node[box,right=.6cm of seed] (add) {Add one\\customer request};
\node[box,right=.6cm of add] (check) {Check policy\\and account state};
\node[box,right=.6cm of check] (pool) {Keep valid\\combinations};
\draw[arr] (seed) -- (add);
\draw[arr] (add) -- (check);
\draw[arr] (check) -- (pool);
\draw[arr] (pool.north) -- ++(0,.45) -| node[pos=.25,above,font=\small]{Extend again, up to four requests} (add.north);
\node[box,below=.7cm of check,text width=5.6cm] (select) {Use development results to estimate difficulty;\\select 100 tasks with diversity limits};
\draw[arr] (pool.south) |- (select.east);
\end{tikzpicture}
\caption{Task construction and selection. The loop checks each added request
against the account state left by earlier requests. Development runs inform
the final selection; not every generated combination was run.}
\label{fig:generation}
\end{figure}

We start with 112 single-request cases. To build a longer call, we append a
request to an existing case and compute the expected result using an
executable model of the utility policy. We keep the new task only if the
request is valid on the account after the preceding actions, adds a distinct
requirement, and does not follow a transfer. Accepted tasks can be extended
again, up to four requests (Figure~\ref{fig:generation}). This produces
2{,}008 eligible combinations.

We then use development runs of a reference agent to estimate difficulty.
Observed success falls from 94/112 for single requests to 29/39, 32/51, and
27/106 for two, three, and four requests. A model fitted to these outcomes
ranks the combinations by predicted success. We select 100 difficult tasks
while limiting repeated requests and accounts. Requests for successful
due-date extensions are excluded because the billing service cannot perform
them on finalized invoices. The final set contains 6 tasks
with two requests, 30 with three, and 64
with four. An independent model reviewed all 100 against their account data
and checks; the identified check defects were corrected. Selection details
are in Appendix~\ref{app:selection}.

For example, a caller asks whether paperless billing is enabled and then asks
to enable it after confirming the destination. Success requires the agent to
verify the caller, report the existing setting, read the email address aloud,
wait for the caller, and change the setting. The checks cover both the spoken
answer and the resulting account change.

\subsection{Callers and background conditions}
\label{sec:caller}

A Gemini Live session (\texttt{gemini-3.1-flash-live-preview}) plays the caller,
using the task's identity, private knowledge, opening line, and requests.
The calm persona is concise and practical. The angry persona is terse,
impatient, and more forceful when misunderstood. Both must follow the same
request order, amounts, consent conditions, and refusals. The angry persona
is not instructed to ask for a supervisor or add demands.

The platform mixes street noise, caf\'e chatter, television speech, or no
background into the caller's outgoing audio. Non-clean conditions use the
platform's noise setting of 0.3\% of a fixed reference speech energy.
Each asset is scaled by its measured RMS; the setting is not an
instantaneous speech-to-noise ratio.
Television uses a tech-news recording with periodic gaps. Each task keeps the same persona across all three repeats, assigned using
a fixed casting seed. Background assignment uses a shuffle seeded by
the trial ID and rotates conditions across repeats. The saved records confirm
identical assignments across all stacks for all 300 task--repeat pairs.
Before exclusions, each stack receives 150 calls per persona and 75 per
background condition. The utterances themselves remain stochastic.

\subsection{Grading}
\label{sec:grading}

Each task has three assertions derived from its account and intended actions:
(1) no account tool use or disclosure before verification, with verification
and human transfer allowed; (2) exactly the expected successful account
changes, in order and with the required values; and (3) the task's remaining
requirements, such as a quoted amount or email confirmation. Spoken values
are checked in the relevant part of the conversation, with stated tolerances.
Repeating a fact supplied by the caller does not count as disclosing a secret.

The LLM-as-a-verifier reads the transcript and recorded tool calls, arguments,
and results. All three assertions must pass. Account changes are inferred
from successful tool results; the verifier does not independently inspect the
backend ledger on this run. The audit uses that ledger as additional evidence.
This distinction matters when a trace loses or misorders a result.

\section{LLM-as-a-verifier}
\label{sec:verifier}

A conversational check can have a precise answer while requiring flexible
language interpretation. The agent might state a payment as ``sixty-five
fifty-two'' or ``sixty-five dollars and fifty-two cents.'' The verifier must
identify the figure, the speaker, its topic, and whether it preceded consent
or action. Our code verifier used regular expressions and a spoken-number
parser. Those rules sometimes confused a postal code with a quantity or a
number of payments with a number of months.

We give the LLM explicit checks specifying the value, tolerance, evidence
window, and relevant topic. It returns a verdict and supporting evidence.
We call this LLM-as-a-verifier to describe the constrained task; it remains
fallible. The main-run assertions also allow semantic equivalents for
statements such as whether an account is past due.

In a separate calibration run of 97 completed
attempts on this exam, we compared code and LLM judgments on the same
744 checks and evidence slices.
\texttt{gpt-5.6-terra} agreed on 99.1\% and
\texttt{gpt-5.6-luna} on 98.9\%.
At the call level, each agreed on 85 of
88 scorable attempts. Review of the
combined disagreements found 7 where the LLM
was right and 8 where the code was right.
All code errors involved speech parsing; LLM errors included missed evidence
and inconsistent verdicts. Agreement measures consistency between two
fallible methods, not accuracy against ground truth.
Appendix~\ref{app:verifier} breaks the comparison down by check.

\section{Evaluation}
\label{sec:metrics}

We ran fourteen stacks three times on each task:
4{,}200 calls, with at most 5 concurrent
attempts and a 900\,s timeout. Six stacks use
speech-to-speech models, three use bundled voice platforms, and five use
assembled speech-recognition, language-model, and speech-synthesis pipelines.
Appendix~\ref{app:systems} lists the configurations. Six share
\texttt{gpt-4.1-mini}; all share Rory's prompt and tools. The stack
implementations are public at \url{https://github.com/veris-ai/rory-agent}.

We exclude 402 calls affected by benchmark
faults, whether they passed or failed, leaving 3{,}798
counted attempts. Agent-side run failures count as misses. Completion is the
fraction of counted attempts passing all assertions. Assertion and transfer
rates, latency, and cost use calls with verdicts and the required records;
Appendix~\ref{app:metrics} gives those denominators. Intervals are Wilson
95\% intervals over each stack's counted attempts. Repeats change the background while retaining each task's persona and
sampling another conversation.

\section{Results}
\label{sec:results}

\begin{table}[htbp]
\centering\small
\caption{Completion, median response latency, and mean cost per
call (Appendix~\ref{app:metrics}). Intervals are Wilson 95\%; $n$ is the number of counted attempts.
Latency and cost use calls with verdicts. Configurations are in
Appendix~\ref{app:systems}.}
\label{tab:main}
\begin{tabular}{@{}lrrrrr@{}}
\toprule
Stack & Complete (\%) & 95\% interval & $n$ & Latency (s) & Cost (\$) \\
\midrule
Grok Voice & 44.7 & [38.8, 50.7] & 264 & 2.64 & 0.210 \\
Gemini 3.8 Live & 40.1 & [34.6, 45.9] & 284 & 1.66 & 0.193 \\
OpenAI GPT-Live 1 & 39.2 & [33.5, 45.3] & 260 & 1.62 & 0.192 \\
OpenAI Realtime 2.1 & 37.9 & [32.3, 43.8] & 269 & 2.12 & 0.633 \\
ElevenLabs & 36.4 & [30.9, 42.2] & 275 & 2.18 & 0.272 \\
OpenAI Realtime & 33.6 & [28.2, 39.5] & 265 & 2.32 & 0.607 \\
LiveKit & 29.4 & [24.5, 34.9] & 289 & 2.96 & 0.077 \\
Gemini 3.1 Live & 25.6 & [20.9, 31.0] & 285 & 2.66 & 0.199 \\
Mistral \textperiodcentered{} Medium 3.5 + Voxtral & 22.7 & [18.1, 28.0] & 269 & 3.75 & 0.279 \\
Gradium & 21.3 & [16.9, 26.6] & 272 & 2.42 & 0.120 \\
Vapi & 20.5 & [15.6, 26.4] & 210 & 2.08 & 0.271 \\
Deepgram & 19.8 & [15.6, 24.8] & 288 & 2.04 & 0.193 \\
Hugging Face \textperiodcentered{} gpt-oss-120b & 19.0 & [14.9, 23.9] & 290 & 5.28 & 0.053 \\
Pipecat \textperiodcentered{} GPT-4.1 mini & 17.3 & [13.3, 22.1] & 278 & 1.96 & 0.109 \\
\bottomrule
\end{tabular}
\end{table}

\paragraph{Completion and account changes.} Completion ranges from
17.3\% to 44.7\%, with a pooled rate
of 29.1\% (Table~\ref{tab:main}). Grok Voice leads, followed
by Gemini 3.8 Live and GPT-Live; their intervals overlap. The three cost about
the same per call (\$0.21,
\$0.193, and
\$0.192). The cheapest stacks are the
Hugging Face, LiveKit, and Pipecat cascades; the two OpenAI Realtime models
are the most expensive, at about three times the leaders' cost. The six stacks sharing \texttt{gpt-4.1-mini} range from
17.3\% to
36.4\%, showing that the language model alone
does not determine performance.

The verification gate passes on 99.0\%
of calls with verdicts, while the required account changes pass on
54.8\%. The gate measures adherence
to the access restriction, not successful authentication: a call that never
verifies can still pass the gate if it discloses nothing and uses no account
tool. On 44.4\% of calls with verdicts,
the gate passes and the account-change assertion fails.

\paragraph{More requests, lower success.} The success rate decreases as
customers make more requests: 56.3\% for two,
34.9\% for three, and
23.8\% for four. Agents sometimes miss a confirmed
action when the caller raises the next request, omit a requested figure, or
state the wrong policy limit. These are different task groups, so the rates
do not isolate the effect of adding a request to an otherwise identical call.

\begin{figure}[t]
\centering
\includegraphics[width=\linewidth]{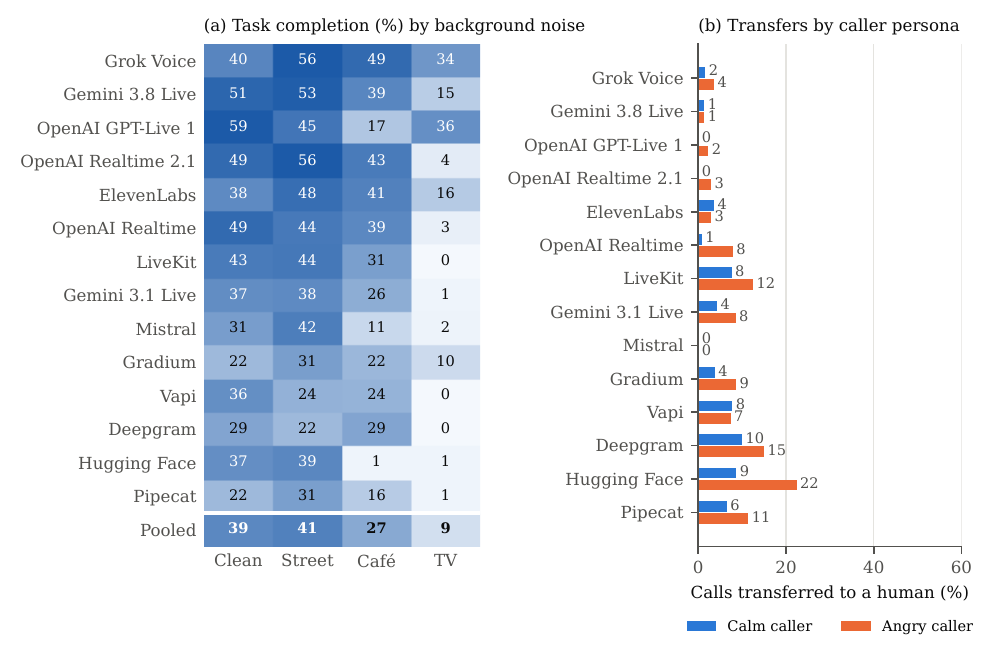}
\caption{(a) Completion by background condition, with
46--75 counted attempts per stack and condition.
(b) Transfers to a human by persona, among calls with verdicts and traces.
All exam tasks are intended to be resolved by the agent.}
\label{fig:noise}
\end{figure}

\paragraph{Background speech.} Pooled completion is
38.7\% in clean audio,
41.4\% with street noise,
27.4\% with caf\'e chatter, and
8.6\% with television
(Figure~\ref{fig:noise}). Grok and GPT-Live retain the highest TV completion
rates; 9 stacks complete fewer than 5\% of
TV calls. Traces show agents treating the broadcast as caller speech,
answering it, or cancelling their own replies. Configuration matters:
our Realtime bridge uses server voice-activity detection without noise
reduction, and our Vapi bridge disables denoising.

\paragraph{Angry callers and transfers.} Angry calls complete at
30.4\%, compared with
27.8\% for calm calls. Persona groups contain
different tasks, so this does not establish that anger improves performance.
Agents transfer 7.6\% of angry calls
and 4.1\% of calm calls to humans.
The persona does not request escalation, and no exam task requires a transfer.

\paragraph{Variation across repeats.} The median stack's completion rate has
a standard deviation of 3.8 percentage points
across the three repeats. For the median stack,
45 tasks have both passing and failing counted
attempts. This combines variation in conversations with the intended changes
in background noise. The overlapping intervals warrant caution about
ranking adjacent stacks.

\paragraph{A satisfied caller can be wrong.}
\label{sec:failures}
The caller declares its goal met on 2{,}072 calls, but
1{,}122 of those fail verification
(54.2\%). The caller hears an assurance
such as ``your payment is recorded'' without seeing whether a tool actually
recorded the right amount. The verifier can compare that assurance with the
tool result. This is why caller satisfaction alone is insufficient to score
these tasks.

\paragraph{Three unsolved tasks.}
\label{sec:solvable}
At least one stack passes 97 tasks. The
remaining three require reporting an overdue arrangement instalment even
though the bill balance is current. The shared account tool reports only
bill arrears, returning zero; agents repeat that figure instead of checking
the arrangement tool. The needed information is available, but without a
successful reference solution we cannot claim to have demonstrated these
three tasks are solvable.

\section{Benchmark validity and limitations}
\label{sec:validity}
\label{sec:limits}

The simulated caller is a voice agent and sometimes departs from its brief:
it may skip a request, supply a figure it should ask for, or hang up while
waiting. We remove 305 affected calls,
60 with confirmed incorrect verdicts,
and 37 lost platform sessions. Of the
402 exclusions,
36 had passed. Removing both passes and
failures avoids retaining calls made artificially easy by the benchmark.
We do not change verdicts or rerun selected calls.

The verdict audit covers 228 calls, using transcripts,
traces, candidate logs, and backend ledgers. Errors included stale interruption
markers, missing tool results, incorrect event order, and misapplied
numeric tolerances. Much of the audit targeted suspected errors; its error
frequency is not an unbiased estimate of verifier accuracy. Exclusions
raise pooled completion from 27.2\% to
29.1\% and preserve the top five stacks' order.
Appendix~\ref{app:audit} reports the per-stack changes. Unaudited verdicts
can still be wrong, and selective exclusions may affect the task mix.

The authors implemented the agent integrations. Their defects count against
the candidate, including noisy-audio settings, cancelled greetings, and the
misleading bill-arrears field. Shared Soundex name matching is lenient and
helps some stacks more than others. Each vendor is represented by one
configuration; vendor tuning could change the results. Costs exclude
telephony and hosting. Cascade costs use published prices and count only
the synthesized characters the caller heard. GPT-Live's cost excludes its
delegated backend model, which would add at most 17\%, and Gemini 3.1 Live's
cost is estimated (Appendix~\ref{app:metrics}).

The task set covers one utility policy and a selected range of modeled
household usage. It is not representative of all utility customers or
regulations. Policy-based admission and model review do not prove that every
task is solvable. Two caller personas and three repeats provide limited
coverage, and noise or persona comparisons remain sensitive to the task mix.

\section{Conclusion}

VAmoS Energy evaluates whether voice agents can complete several customer
requests while following policy, using stateful services, and handling
background speech. Across fourteen configurations,
completion ranges from 17.3\% to
44.7\%. Television disrupts most stacks, account changes
fail more often than the verification gate, and callers often accept
incorrect outcomes. Evaluating both conversation and execution exposes
failures that a caller's apparent satisfaction would miss.

\begin{ack}
All authors are affiliated with Veris AI, which developed the simulation
platform, Stripe twin, and benchmark infrastructure. Usage profiles are from
NREL ResStock 2022.1 (CC BY 4.0), aggregated and selected by the authors.
The television background uses Daily Tech News Show audio (CC BY).
\end{ack}

\clearpage
\bibliographystyle{plainnat}
\begingroup\raggedright
\bibliography{references}
\endgroup
\clearpage
\appendix

\section{Policy encoded in the benchmark}
\label{app:policy}

The system prompt caps arrangements at 60, 36, 12, or 6 months for income
bands at or below 150\%, 151--250\%, 251--300\%, and above 300\% of the
federal poverty level. Missing income uses a six-month cap. These are the
benchmark's encoded limits. Chapter~56 governs payment arrangements but the
cited chapter alone does not establish this exact cap table; the mapping
needs separate legal validation. The benchmark also encodes winter
termination protection from December~1 to March~31 for recorded income at or
below 250\%, and medical protection through the recorded certificate date.
Chapter~56 includes qualifications and exceptions that the simplified task
policy does not fully model \citep{pacode56}.

Authored business rules require at least \$50 in arrears and a two-month
minimum arrangement, a 25\% down payment after a broken arrangement within
twelve months, and at most one extension of 1--15 days per rolling year.
Accounts with at least \$500 outstanding and 60 days past due must be
transferred to a person. The final exam excludes tasks requiring that
transfer or a successful extension. These assumptions define the benchmark;
they are not presented as statutory requirements.

\section{Task selection}
\label{app:selection}

A combination cannot repeat a request, continue after a transfer, or add a
request whose account conditions no longer hold. A refusal must also be
valid on the initially seeded account, so an earlier action cannot manufacture
the reason for refusal. The complete task must require more than either the
preceding requests or the final request alone. The 2{,}008 admitted tasks
contain requests from 70 of the 112 starting cases.

Difficulty estimation fits one success probability per starting request by
maximum likelihood on development outcomes, and predicts a combined task's
success as the product of its constituent probabilities. The 100-task
selection allows one ordering per set of requests, at most five tasks per
account, and at most 25 uses of a request except the common balance and
paperless requests. The selected set draws on 26 starting
requests across 24 accounts. Its predicted mean success
was 0.27. This model ranks tasks for selection; it is not a calibrated
forecast of performance across all evaluated stacks.

\section{Verifier comparison}
\label{app:verifier}

The calibration comparison used 97 completed attempts from September~14,
2026, before the main September~24--28 run and its semantic-assertion update.
Only 88 had sufficient evidence for a
complete code verdict. Per-check comparisons include the checks that can be
scored in the remaining attempts. Table~\ref{tab:verifier} reports agreement.
The code verifier is a comparator, not a source of ground-truth labels.

\begin{table}[H]
\centering
\small
\caption{Agreement between the code verifier and two LLM verifiers, check by
check, on 97 completed attempts from a calibration run of this exam. ``Reads''
is the evidence a check needs; the speech rows are the ones where the code
extracts values with patterns.}
\label{tab:verifier}
\begin{tabular}{@{}llrrr@{}}
\toprule
Check & Reads & Checks & \texttt{terra} agrees & \texttt{luna} agrees \\
\midrule
Exact world effects, in order & tool results & 97 & 96 & 96 \\
No account tool before verification & tool calls & 97 & 97 & 97 \\
Required tool called & tool calls & 20 & 20 & 20 \\
Forbidden tool not called & tool calls & 34 & 34 & 34 \\
Quote, caller turn, then create & tool calls + turns & 34 & 32 & 33 \\
No secret spoken before verification & speech & 97 & 95 & 95 \\
Figure stated in a clause on the topic & speech & 224 & 223 & 221 \\
One of a set of phrases used & speech & 130 & 130 & 130 \\
One of several figures stated & speech & 6 & 5 & 5 \\
Figure relayed from a tool result & speech + tool results & 5 & 5 & 5 \\
\midrule
All checks & & 744 & 737 (99.1\%) & 736 (98.9\%) \\
\bottomrule
\end{tabular}
\end{table}

The prose-assertion judge agreed with code on
76 of
88 scorable calls. Asking explicit checks
with matched evidence raised agreement to 85.
Both the specification and evidence presentation changed, so this comparison
does not isolate wording alone. Code errors included misreading spoken
postal codes and confusing instalment counts with term lengths. LLM errors
included missing a disable-then-enable pair, overlooking a clause, and
returning a verdict inconsistent with quoted evidence.

\section{Additional metrics}
\label{app:metrics}

Completion uses all counted attempts, including agent-side run failures.
Assertion rates use the 3{,}635 calls with verdicts; transfer
rates use calls with verdicts and traces. A transfer is a call to
\texttt{transfer\_to\_human}, not a spoken promise to transfer. Response
latency is the interval from caller speech ending to agent speech starting,
pooled over agent turns. Cost is per call with a verdict. For the cascades it is
each vendor's published price applied to measured usage: audio minutes,
call minutes, and synthesized characters. The speech-to-speech models from
OpenAI and Google bill every reply for everything in the session so far,
including the prompt, tool definitions, and earlier audio, so we take their
cost from the vendors' bills. Each speech-to-speech stack ran all its calls on
one day, and no other trial used the same model that day; we spread each bill
over the stack's calls in proportion to call length. Gemini 3.1 Live uses the
same model as the simulated caller, so its bill cannot be separated; we scale
it by Gemini 3.8 Live's ratio of bill to estimate, since the two share prices
and a bridge. Cost per completed call is the total cost over passes. It is
lowest for the Hugging Face and LiveKit cascades
(\$0.22 and
\$0.26), about
\$0.47 for the three leading stacks,
and highest for the two OpenAI Realtime models
(\$1.62 and
\$1.72).

Table~\ref{tab:cuts} gives completion by request count, persona, and the
last request's use case. A call must satisfy all its requests, so the last
request's use case need not identify the source of failure.
\begin{table}[H]
\centering
\small
\caption{Completion (\%) by number of requests, caller persona, and use case of
the last request. Before exclusions, each stack has 18/90/192 attempts with 2/3/4 requests,
150 per persona, and 111/189 per use case; counted denominators are smaller.}
\label{tab:cuts}
\setlength{\tabcolsep}{4pt}
\begin{tabular}{@{}lccccccc@{}}
\toprule
 & \multicolumn{3}{c}{Requests} & \multicolumn{2}{c}{Persona} & \multicolumn{2}{c}{Use case} \\
\cmidrule(lr){2-4}\cmidrule(lr){5-6}\cmidrule(lr){7-8}
Stack & 2 & 3 & 4 & Calm & Angry & High bill & Payment \\
\midrule
Grok Voice & 81 & 46 & 40 & 39 & 50 & 40 & 48 \\
Gemini 3.8 Live & 56 & 49 & 34 & 34 & 46 & 31 & 45 \\
OpenAI GPT-Live 1 & 65 & 45 & 34 & 44 & 34 & 38 & 40 \\
OpenAI Realtime 2.1 & 67 & 51 & 28 & 34 & 41 & 26 & 45 \\
ElevenLabs & 71 & 38 & 32 & 31 & 41 & 35 & 37 \\
OpenAI Realtime & 62 & 37 & 29 & 34 & 33 & 27 & 38 \\
LiveKit & 59 & 41 & 21 & 30 & 29 & 22 & 34 \\
Gemini 3.1 Live & 56 & 35 & 18 & 26 & 25 & 19 & 29 \\
Mistral \textperiodcentered{} Medium 3.5 + Voxtral & 44 & 26 & 19 & 20 & 26 & 22 & 23 \\
Gradium & 71 & 24 & 15 & 20 & 23 & 21 & 21 \\
Vapi & 18 & 18 & 22 & 20 & 21 & 10 & 27 \\
Deepgram & 41 & 24 & 16 & 23 & 17 & 18 & 21 \\
Hugging Face \textperiodcentered{} gpt-oss-120b & 56 & 24 & 14 & 19 & 19 & 16 & 20 \\
Pipecat \textperiodcentered{} GPT-4.1 mini & 29 & 25 & 12 & 15 & 19 & 15 & 19 \\
\midrule
Pooled & \textbf{56.3} & \textbf{34.9} & \textbf{23.8} & \textbf{27.8} & \textbf{30.4} & \textbf{24.6} & \textbf{31.8} \\
\bottomrule
\end{tabular}
\end{table}

\begin{figure}[ht]
\centering
\includegraphics[width=\linewidth]{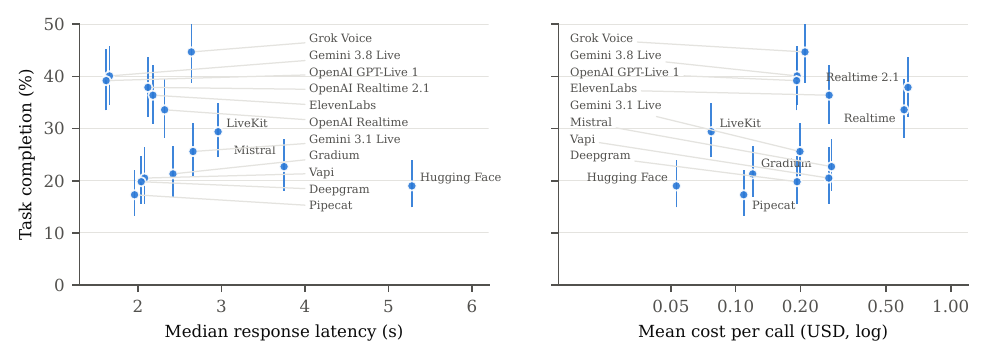}
\caption{Completion against median response latency and mean cost per
call. Vertical bars are Wilson 95\% intervals; the cost axis is
logarithmic.}
\label{fig:tradeoffs}
\end{figure}

\section{Audit and exclusions}
\label{app:audit}

A stratified audit of 72 calls, three passes
and three failures for each of twelve initially available stacks, found no
wrong verdict. Sampling favored calls with possible evidence problems. A targeted sweep of
116 suspected errors found
43 false failures and
1 false pass. Another
40 calls yielded
1 false failure. These three groups cover 228 calls. A
separate artifact-by-artifact review of 28
calls found 27
verdicts correct.

The most common false-failure cause was a stale interruption marker
(17 calls), followed by
incorrect tolerance handling
(13). Other causes included
overlooked evidence, literal wording judgments, missing real tool results,
and tool results timestamped before their calls. The audit used candidate
logs and backend ledgers to distinguish trace defects from failed actions.

The largest caller deviation affected Vapi: filler phrases while a tool ran
caused the simulated caller to hang up despite instructions to wait.
Caller-side exclusions remove
85 Vapi calls. Such deviations
can also make a task easier by omitting a request or supplying an answer;
therefore the exclusion list includes passes. Table~\ref{tab:validity}
reports all exclusions and both completion rates.
\begin{table}[H]
\centering
\scriptsize
\caption{Completion over all attempts and over counted attempts, with the
attempts excluded and why: the simulated caller broke its script
(``caller''), the verifier misjudged the call (``verifier''), or the platform
lost the session (``lost''). ``Passes'' is the number of excluded attempts
that had passed. ``Soundex'' is the number of passes, over all attempts,
whose caller was verified only through the sound-alike name match
(described below the table). Ordered by counted completion.}
\label{tab:validity}
\setlength{\tabcolsep}{3.2pt}
\begin{tabular}{@{}lcccccccccc@{}}
\toprule
Stack & All (\%) & Rank & Counted (\%) & Rank & Excluded & Caller & Verifier & Lost & Passes & Soundex \\
\midrule
Grok Voice & 41.0 & 1 & 44.7 & 1 & 36 & 25 & 11 & 0 & 5 & 43 \\
Gemini 3.8 Live & 38.0 & 2 & 40.1 & 2 & 16 & 11 & 2 & 3 & 0 & 25 \\
OpenAI GPT-Live 1 & 37.3 & 3 & 39.2 & 3 & 40 & 30 & 10 & 0 & 10 & 26 \\
OpenAI Realtime 2.1 & 35.0 & 4 & 37.9 & 4 & 31 & 23 & 6 & 2 & 3 & 28 \\
ElevenLabs & 34.3 & 5 & 36.4 & 5 & 25 & 13 & 4 & 8 & 3 & 28 \\
OpenAI Realtime & 30.3 & 6 & 33.6 & 6 & 35 & 31 & 4 & 0 & 2 & 25 \\
LiveKit & 28.7 & 7 & 29.4 & 7 & 11 & 6 & 5 & 0 & 1 & 32 \\
Gemini 3.1 Live & 24.7 & 8 & 25.6 & 8 & 15 & 14 & 1 & 0 & 1 & 16 \\
Mistral \textperiodcentered{} Medium 3.5 + Voxtral & 22.0 & 9 & 22.7 & 9 & 31 & 9 & 2 & 20 & 5 & 17 \\
Gradium & 19.3 & 11 & 21.3 & 10 & 28 & 27 & 1 & 0 & 0 & 14 \\
Vapi & 14.7 & 14 & 20.5 & 11 & 90 & 85 & 2 & 3 & 1 & 14 \\
Deepgram & 19.7 & 10 & 19.8 & 12 & 12 & 11 & 1 & 0 & 2 & 27 \\
Hugging Face \textperiodcentered{} gpt-oss-120b & 18.7 & 12 & 19.0 & 13 & 10 & 7 & 2 & 1 & 1 & 21 \\
Pipecat \textperiodcentered{} GPT-4.1 mini & 16.7 & 13 & 17.3 & 14 & 22 & 13 & 9 & 0 & 2 & 20 \\
\bottomrule
\end{tabular}
\end{table}

The shared verification tool uses Soundex for names, retaining the first
letter while allowing some spelling differences. It can accept a sound-alike
name but reject ``Kamila'' for ``Camila.'' Across all attempts, including
those later excluded, 1{,}027 calls verified only
through sound-alike matching and 336 passing calls
depend on it. These counts describe the original run and use raw denominators.
All stacks use the same matcher, but its benefit differs with their
transcriptions.

\section{Voice stacks}
\label{app:systems}

\begin{table}[H]
\centering
\small
\caption{The fourteen stacks in this run, in order of
observed completion. Every stack runs the same prompt, tool declarations, and
dispatcher from the shared core. Models are as recorded in the candidate logs
and the implementations' configuration.}
\label{tab:systems}
\begin{tabular}{@{}ll>{\raggedright\arraybackslash}p{7.2cm}@{}}
\toprule
Stack & Class & Models (as run) \\
\midrule
Grok Voice & Speech-to-speech & \texttt{grok-voice-think-fast-2.0}, voice \texttt{eve} \\
Gemini 3.8 Live & Speech-to-speech & \texttt{gemini-3.8-live}, same bridge as Gemini 3.1 Live \\
OpenAI GPT-Live 1 & Speech-to-speech & \texttt{gpt-live-1}, tool use delegated to \texttt{gpt-5.6-terra}, voice \texttt{marin} \\
OpenAI Realtime 2.1 & Speech-to-speech & \texttt{gpt-realtime-2.1}, same bridge as OpenAI Realtime \\
ElevenLabs & Bundled platform & Conversational AI: ElevenLabs ASR $\to$ \texttt{gpt-4.1-mini} $\to$ \texttt{eleven\_flash\_v2} \\
OpenAI Realtime & Speech-to-speech & \texttt{gpt-realtime-2}, voice \texttt{alloy}, server VAD (threshold 0.5) \\
LiveKit & Assembled cascade & LiveKit Agents: Deepgram \texttt{nova-3} $\to$ \texttt{gpt-4.1-mini} $\to$ ElevenLabs \texttt{eleven\_flash\_v2}, Silero VAD \\
Gemini 3.1 Live & Speech-to-speech & \texttt{gemini-3.1-flash-live-preview} \\
Mistral & Assembled cascade & Voxtral realtime STT $\to$ \texttt{mistral-medium-2604} $\to$ Voxtral TTS, Silero VAD \\
Gradium & Assembled cascade & gradbot engine: Gradium STT $\to$ \texttt{gpt-4.1-mini} $\to$ Gradium TTS \\
Vapi & Bundled platform & Hosted orchestrator: Deepgram $\to$ \texttt{gpt-4.1-mini} $\to$ ElevenLabs; tools as server tools \\
Deepgram & Bundled platform & Voice Agent API: \texttt{nova-3} $\to$ \texttt{gpt-4.1-mini} $\to$ Aura-2 \\
Hugging Face & Assembled cascade & Whisper large-v3 $\to$ \texttt{gpt-oss-120b} (via the Hugging Face router) $\to$ Kokoro-82M, Silero VAD \\
Pipecat & Assembled cascade & Pipecat: Deepgram \texttt{nova-3} $\to$ \texttt{gpt-4.1-mini} $\to$ ElevenLabs \texttt{eleven\_flash\_v2} \\
\bottomrule
\end{tabular}
\end{table}

\section{Tools}
\label{app:tools}

\begingroup\raggedright
\texttt{verify\_caller}, \texttt{get\_account}, \texttt{list\_bills},
\texttt{get\_bill}, \texttt{explain\_bill}, \texttt{compare\_bills},
\texttt{get\_payment\_history}, \texttt{make\_payment},
\texttt{quote\_payment\_arrangement}, \texttt{create\_payment\_arrangement},
\texttt{get\_payment\_arrangement}, \texttt{modify\_payment\_arrangement},
\texttt{request\_due\_date\_extension}, \texttt{set\_paperless\_billing},
\texttt{get\_supplier\_info}, \texttt{transfer\_to\_human}. The five
world-changing tools are \texttt{make\_payment},
\texttt{create\_payment\_arrangement}, \texttt{modify\_payment\_arrangement},
\texttt{request\_due\_date\_extension}, and \texttt{set\_paperless\_billing}.

\par\endgroup

\section{Requests in the exam}
\label{app:asks}

\begin{table}[H]
\centering
\small
\caption{The requests the 100 tasks draw on, with the number
of tasks that include each.}
\label{tab:asks}
\begin{tabular}{@{}lr@{}}
\toprule
Request & Tasks \\
\midrule
Enable paperless after confirming the email & 73 \\
Report the current balance and delinquency status & 50 \\
Reject a one-month arrangement & 25 \\
Record the caller's exact partial instalment & 25 \\
Record a payment made today & 25 \\
Quote before creating an arrangement & 22 \\
Redirect a second-plan request to the active plan & 21 \\
Create an eligible arrangement after informed consent & 14 \\
Refuse a second concurrent arrangement & 13 \\
Do not silently shorten an over-limit request & 8 \\
Reject a zero-day extension & 8 \\
Refuse a sixteen-day extension & 7 \\
Refuse a second extension within twelve months & 5 \\
Cap the term by income tier (four income groups: $\le$150\%, 151--250\%, 251--300\%, $>$300\%) & 20 \\
Do not assume an income tier that is not on record & 5 \\
Reject a thirteen-month arrangement & 5 \\
Do not record a future promise as a payment & 5 \\
Reject an explicitly future-dated instalment & 5 \\
Report active arrangement status & 5 \\
Identify an overdue arrangement instalment & 5 \\
Explain the existing instalment schedule & 5 \\
Distinguish bill payments from plan instalments & 5 \\
Report current paperless status & 2 \\
\bottomrule
\end{tabular}
\end{table}

\end{document}